\documentclass[runningheads]{llncs}
\usepackage{graphicx}

\usepackage{pagecolor}
\usepackage{lipsum}  

\usepackage{url}
\usepackage{algorithm}
\usepackage{algorithmic}
\usepackage{enumitem}
\usepackage{tikz}
\definecolor{light_gray}{HTML}{FFFFFF} 
\definecolor{solid_gray}{HTML}{000000} 
\usepackage{multirow}
\usepackage{array}
\usepackage{hhline}
\usepackage{pifont}
\usepackage{makecell}
\usepackage{array} 
\usepackage{booktabs}
\usepackage{siunitx}
\usepackage{nameref}
\usepackage{xcolor}
\usepackage{transparent}
\usepackage{soul}
\usetikzlibrary{calc}
\usepackage{tcolorbox}
\usepackage{amssymb}
\usepackage{bbding}
\usepackage{pifont}
\usepackage{longtable}
\usepackage{tabularx}
\usepackage{rotating}
\usepackage{adjustbox}
\usepackage{bibentry}
\usepackage{pgfplots}
\usepackage{amsmath}
\usepackage{pgfplotstable}
\usepackage{comment}
\usepackage{float} 
\usetikzlibrary{positioning} 

\usepackage{tikz}
\usetikzlibrary{shapes}

\newtcbox{\highlight}[1][]{%
    colback=green!15!white,
    colframe=green!15!white,
    boxrule=0pt,
    boxsep=0pt,
    left=2pt,
    right=2pt,
    top=2pt,
    bottom=2pt,
    sharp corners,
    #1
}

\definecolor{transblue}{rgb}{0.69, 0.85, 0.96} 
\definecolor{editcolor}{rgb}{0.7, 0, 0} 

\newtcbox{\bleuhl}[1][]{%
    on line,
    arc=0pt,
    outer arc=0pt,
    colback=transblue!50,
    boxsep=0pt,
    left=2.0pt,
    right=2.0pt,
    top=1.0pt,
    bottom=0.4pt,
    boxrule=0.0pt,
    #1
}

\begin{document}
\title{Large Language Models As MOOCs Graders}
\titlerunning{LLMs As MOOCs Graders}
%

\author{Shahriar Golchin\inst{1} \and
Nikhil Garuda\inst{2} \and
Christopher Impey\inst{2} \and
Matthew Wenger\inst{2}}
\authorrunning{S. Golchin et al.}

%

\institute{University of Arizona, Department of Computer Science, Tucson, AZ, USA \and
University of Arizona, Department of Astronomy, Tucson, AZ, USA}

%
\maketitle              
\begin{abstract}
Massive open online courses (MOOCs) unlock the doors to free education for anyone around the globe with access to a computer and the internet. Despite this democratization of learning, the massive enrollment in these courses means it is almost impossible for one instructor to assess every student’s writing assignment. As a result, peer grading, often guided by a straightforward rubric, is the method of choice. While convenient, peer grading often falls short in terms of reliability and validity. In this study, using 18 distinct settings, we explore the feasibility of leveraging large language models (LLMs) to replace peer grading in MOOCs.
Specifically, we focus on two state-of-the-art LLMs: GPT-4 and GPT-3.5, across three distinct courses: Introductory Astronomy, Astrobiology, and the History and Philosophy of Astronomy. To instruct LLMs, we use three different prompts based on a variant of the zero-shot chain-of-thought (Zero-shot-CoT) prompting technique: Zero-shot-CoT combined with instructor-provided correct answers; Zero-shot-CoT in conjunction with both instructor-formulated answers and rubrics; and Zero-shot-CoT with instructor-offered correct answers and LLM-generated rubrics. 
Our results show that Zero-shot-CoT, when integrated with instructor-provided answers and rubrics, produces grades that are more aligned with those assigned by instructors compared to peer grading. However, the History and Philosophy of Astronomy course proves to be more challenging in terms of grading as opposed to other courses. Finally, our study reveals a promising direction for automating grading systems for MOOCs, especially in subjects with well-defined rubrics.


\end{abstract}
\section{Introduction}
Massive open online courses (MOOCs)\footnote{\url{https://www.mooc.org/}} have revolutionized access to education by offering a wide array of free courses to anyone with an internet connection. This platform has democratized education and made it globally accessible, accommodating an unlimited number of participants \cite{impey2016bringing,Impey_Wenger_Austin_2015}.
However, the resulting increase in participant numbers, though beneficial to a broader audience, introduces challenges in delivering personalized and constructive feedback on student performance, including the task of grading their assignments. To address this issue, peer grading has become a common practice. While peer grading enhances student engagement and motivation, its reliability and validity are often questioned \cite{FORMANEK2017243}.

From the application of statistical methods \cite[inter alia]{10.1145/3563357.3567751,10129378,tavabi2022building} to the recent emergence of large language models (LLMs) in the realm of Natural Language Processing, this field has significantly fertilized various other disciplines \cite[inter alia]{DBLP:journals/corr/abs-2303-12712,DBLP:journals/corr/abs-2302-04023,SHAMSHIRI2024105200,DBLP:journals/npjdm/TavabiRSGSHK23} and has also paved the way for addressing more complex tasks \cite{bubeck2023sparks}.
In this study, we investigate the potential of LLMs in handling the task of grading assignments in MOOCs and the viability of replacing the current peer grading system with LLMs on this platform, aiming to establish a more streamlined and automated grading mechanism.
This potentially reduces human intervention in the grading process, thereby serving more students with more accurate feedback and enhancing the overall learning experience.

In this study, we employ three distinct prompting strategies to assess the grading abilities of two state-of-the-art LLMs—GPT-4 \cite{DBLP:journals/corr/abs-2303-08774} and GPT-3.5 \cite{DBLP:conf/nips/Ouyang0JAWMZASR22}—in three MOOC subjects: Introductory Astronomy, Astrobiology, and the History and Philosophy of Astronomy \cite{astrobiology,astronomy,HPA}.
Our methodology is primarily grounded on utilizing a modified version of the zero-shot chain-of-thought prompting (Zero-shot-CoT) technique \cite{NEURIPS2022_8bb0d291}, but we tweak this prompt by providing it with a more in-context understanding of the given task.
In particular, we augment this prompt in a threefold manner: (1) we fortify Zero-shot-CoT with instructor-provided answers; (2) we boost Zero-shot-CoT by integrating both instructor-guided answers and grading rubrics; and (3) we enrich Zero-shot-CoT with correct answers offered by the instructor but supplemented with LLM-generated rubrics based on the correct answers. Our findings suggest that GPT-4 using Zero-shot-CoT prompting with both instructor-provided answers and rubrics outperforms GPT-3.5 and peer grading in terms of score alignment with those given by instructors.
However, grading courses that require imaginative or speculative thinking, such as History and Philosophy of Astronomy, is found to be a challenge for both LLMs and peer grading. Despite this, GPT-4 still excels over peer grading by producing scores more closely aligned with those given by the instructors in such scenarios.

The main contributions of our study are:

\noindent \textbf{(1)} We explore the potential of replacing the peer grading mechanism with the use of LLMs in MOOCs, aiming for an automated and more accurate grading process that minimizes human involvement, particularly in scenarios needing grading feedback for a large number of participants.

\noindent \textbf{(2)} Our experiments encompass 18 distinct scenarios, applying two LLMs (GPT-4 and GPT-3.5) with three prompting techniques (Zero-shot-CoT plus correct answers, Zero-shot-CoT complemented with correct answers and rubrics, and Zero-shot-CoT with correct answers and LLM-presented rubrics) to real-world data from three MOOC topics (Introductory Astronomy, Astrobiology, and the History and Philosophy of Astronomy).

\noindent \textbf{(3)} When integrating Zero-shot-CoT with instructor-crafted correct answers and rubrics, together with GPT-4, it outperforms all other settings, including peer grading, in generating grades more aligned with those allocated by instructors. However, our findings demonstrate that this alignment is more pronounced in courses that require less creative and imaginative thinking.

\section{Related Work}

The existing foundation for this research stems from preceding studies that explored peer review grading in MOOCs \cite{FORMANEK2017243,Gamage_2021}. These studies showed a connection between peer review participation and elements such as student engagement and course completion rates. They also revealed issues related to inconsistency in grades and problems concerning reliability and validity within this process.

Regarding the exploration of LLMs in educational settings, the primary focus has been on their utilization by instructors and students to generate and respond to educational content \cite{Kasneci_2023,Alseddiqi_2023}. One such exemplary application is at Khan Academy, GPT-4 is employed as a learning assistance tool, aiding in student learning via content-specific question-answering \cite{OpenAI_2023}.

Despite its relevance, there is a noticeable gap in research that directly parallels our study. Several studies \cite{Morris_2023,Hiroaki_2022,Riordan_2017} employed the classic encoder-only component of Transformer network \cite{devlin2018bert,vaswani2017attention}, examining the grading of students' assignments within short question-answering contexts. Despite their value, these works are confined to short question-answering settings and necessitate the training of individual, subject-specific models, all the while involving human interaction. The research bearing the closest resemblance to our current endeavor is \cite{Pinto_2023}, wherein GPT-3.5 (integrated within the framework of ChatGPT) is used in a corporate context, providing feedback on employee performance metrics.

Our work, while centers around grading students' assignments, is substantially different from others in the sense that we are the first to use GPT-4 in addition GPT-3.5 to study the possibility of replacing peer grading system in MOOCs, eliminating the necessity for human involvement in the grading process.

\section{Approach}
\subsection{Prompts}
To guide LLMs in grading assignments, we formulate three distinct prompts based on a variation of Zero-shot-CoT.\footnote{We refer to this as a variation of Zero-shot-CoT since this version contains more information compared to the original Zero-shot-CoT, presented in \cite{NEURIPS2022_8bb0d291}.} After conducting a series of preliminary experiments, we found that the Zero-shot-CoT method yields scores more closely aligned with those given by instructors in contrast to vanilla zero-shot prompting. As a result, we chose to employ this prompting approach throughout our study.
Further, this methodology enables us to observe the underlying reasoning for each score generated by the LLM, as the LLM is tasked with speaking out its thoughts, thereby proactively mitigating potential hallucinations during the grade generation process by the LLMs.
We develop three prompts, which incorporate Zero-shot-CoT in combination with: (1) instructor-provided correct answers, (2) instructor-presented answers and rubrics, and (3) correct answers furnished by the instructor, along with the LLM-generated rubric. In the following, we provide a detailed explanation of each prompt.

\noindent \textbf{Zero-shot-CoT with Instructor-provided Answers.} Initially, we implement Zero-shot-CoT by incorporating the correct answers, supplied by course instructors, as in-context information when prompting LLMs to generate grades for assignments. Further, each student's assignment is embedded into the input prompt, and the LLM is tasked with grading based on the instructor-provided answer for each question.
It is important to note that within this prompt, there is no guidance on how the grading should be carried out, and the only information about the grading is the total points available for each question. Finally, the grading is conducted for each student and question separately, ensuring that there is no influence from other questions or the grades of other students on the downstream grades generated by the LLM. Figure \ref{figure:zero-shot-cot-with-answers} illustrates the Zero-shot-CoT prompt with correct answers integrated.



\begin{figure*}[!]
    \begin{minipage}{\textwidth} 
        \centering
        \begin{tikzpicture}[rounded corners=8pt, thick, text=black, text opacity=1]
            \node[draw=solid_gray, fill=light_gray, line width=1pt, 
            text=black, text width=0.97\textwidth, align=left, 
            font=\fontsize{6pt}{1em}\selectfont, inner xsep=5pt, 
            inner ysep=5pt] at (0,0) {\textbf{Instruction:}
            You are a fair and knowledgeable instructor whose task is to evaluate the 
            student's assignments in accordance with the correct answers to each of the 
            questions that are presented in the section that follows.
            Make sure you speak your thoughts aloud so that the students can understand 
            the rationale for any points deducted.
            Let's grade assignments step by step.

            - - -

            \textbf{Questions/Answers:}

            QUESTION 1

            Clearly identify the detection methods used to gather data for each 
            exoplanet.  Briefly explain how each detection method works. 
            Correctly identify both detection methods. Clearly explain how each 
            detection method works. Correctly identify which physical characteristics can be 
            learned from each set of data, and explain why. Clearly identify 
            physical characteristics for both exoplanets. Clearly identify one 
            exoplanet as Earth-like.

            ANSWER 1

            The radial velocity technique was used to gather data to create the 
            radial velocity versus time graphs for Star A and Star B. The radial 
            velocity technique measures the Doppler shift of the star’s absorption 
            lines that are caused by the exoplanet tugging on the star. Using the 
            transit technique would allow you to obtain a stellar brightness versus 
            time graph, like the graph for Star A. Transits take place when the 
            exoplanet passes in front of its star. There is a noticeable dip or 
            drop in the brightness of the star, caused by the exoplanet blocking a 
            bit of the star’s light.

            Using the radial velocity curves, we can determine both the orbital 
            period of exoplanets around their stars as well as the mass of the 
            exoplanet. For Star A, the period is about 200 days. For Star B, the 
            period is about 6 days. As the planet tugs the star toward and away 
            from the observer, we can tell how long it takes to go around the star
            and how hard it pulls on the star. We can infer from the periods 
            that Exoplanet B is closer to its star than Exoplanet A is to its 
            star. The larger radial velocity means that Exoplanet B is massive 
            and very close to Star B. Exoplanet B is likely a Hot Jupiter based
            on this data. Using the transit data, you can estimate the radius or 
            size of the exoplanet. The ratio of exoplanet radius to star radius 
            is about 0.014. This means one exoplanet radius is about 0.014 of a 
            star radius. The Earth is about 0.009 of the Sun’s radius, creating
            a comparable ratio between Exoplanet A-Star A and the Earth-Sun.

            Based on the data, Star A is the more likely candidate for an 
            Earth-like and habitable exoplanet. It has a period shorter than 
            Earth’s, a distance shorter than the Earth-Sun distance, and has 
            a larger planet to star ratio than the Earth to the Sun. In contrast, 
            Star B has a dramatically smaller period, and Exoplanet B is much 
            closer to Star B, meaning it is not likely to harbor life.
            
            \textbf{Grade: 10/10}

            - - -

            \textbf{Student's Answers:}

            The methods used to collect data from the exoplanets were the transit and radial velocity. 
            The transit method works through the monitoring of the starlight of a particular star for periodic dimming since that is an indication that an exoplanet is blocking out a portion of that starlight as it orbits, to apply this method the orbital plane needs to be tilted towards the observer. 
            The radial velocity method measures the reflex motion of stars to detect exoplanets, because the absorbed light is periodically red and then blue-shifted recurring each orbit, the data can be added together. This measurement is done by taking several spectra of the star and comparing them. [...]
            
            \textit{\{The answer is partially cut due to space constraints.\}}

            \textbf{Grade: }

            - - -

            \highlight{
                9/10
                }
            \vspace{-0.30cm}
            \highlight{
                The student has correctly identified the detection methods 
                used to gather data for each exoplanet and
                }
            \vspace{-0.30cm}
            \highlight{explained how 
            identified which physical characteristics can be learned from 
            each detection method
            }
            \vspace{-0.30cm}
            \highlight{works. The student has also correctly 
            each set of data and explained why. However, the student made a}
            \vspace{-0.30cm}
            \highlight{small mistake in the analysis of the radial velocity graphs. The student stated that the radial velocity
            }
            \vspace{-0.30cm}
            \highlight{of Star A lasts for nearly 275 days and Star B lasts for 11 days, which is incorrect. The correct}
            \vspace{-0.30cm}
            \highlight{interpretation should be that the orbital period of Exoplanet A is about 275 days and Exoplanet B is}
            \vspace{-0.30cm}
            \highlight{about 11 days. The student correctly identified Exoplanet A as more Earth-like. Overall, the student}
            \vspace{-0.30cm}
            \highlight{demonstrated a good understanding of the topic but made a minor error in interpreting the data.}
            };
        \end{tikzpicture}
    \end{minipage}
    \caption{An illustration of the Zero-shot-CoT prompt along with answers provided by the course instructor to grade students' assignments. Each question is assessed individually for every student. We repeat this process for all questions and students, incorporating their answers into the prompt, and tasking the LLM to grade their assignments.}
    \label{figure:zero-shot-cot-with-answers}
\end{figure*}

\begin{figure*}[!]
    \begin{minipage}{\textwidth} 
        \centering
        \begin{tikzpicture}[rounded corners=8pt, thick, text=black, text opacity=1]
            \node[draw=solid_gray, fill=light_gray, line width=1pt, 
            text=black, text width=0.97\textwidth, align=left, 
            font=\fontsize{6pt}{1em}\selectfont, inner xsep=5pt, 
            inner ysep=5pt] at (0,0) {\textbf{Instruction:}
            You are a fair and knowledgeable instructor whose task is to 
            evaluate the student's assignments in accordance with the correct 
            answers and provided rubric to each of the questions that are 
            presented in the section that follows.
            
            Make sure that the score deduction follows the provided rubric.
            Let's grade assignments step by step and explain the reasons behind the point 
            deduction given the rubric.

            - - -

            \textbf{Rubric:}

            QUESTION 1

            Correctly identify which physical characteristics can be learned 
            from each set of data, and explain why.

            2 points: The writer clearly identifies physical characteristics 
            for both exoplanets

            1 point: The writer identifies physical characteristics of only 
            one exoplanet

            0 points: The writer does not clearly identify physical 
            characteristics of either exoplanet

            Clearly identify the detection methods used to gather data for 
            each exoplanet.  Briefly explain how each detection method works.

            4 points: The writer correctly identifies both detection methods; 
            clear explanation of how each detection method works

            3 points: The writer correctly identifies both detection methods; 
            explanation for one or both methods is unclear or incomplete

            2 points: The writer correctly identifies and explains one 
            detection method; does not identify other detection method, 
            no explanation

            1 point: The writer correctly identifies one detection method; 
            explanation of detection method unclear or incomplete

            0 points: The writer does not identify or explain either 
            detection method

            Clearly identify one exoplanet as Earth-like
            
            1 point: Writer clearly identifies one exoplanet as Earth-like

            0 points: Writer does not clearly identify one exoplanet as 
            Earth-like

            Student writes with sufficient clarity and detail to communicate 
            their points effectively

            2 points: Writer presents arguments in clear, logical way that 
            demonstrates understanding of concepts; correct use of 
            scientific terms/language.  Connections between concepts are 
            well developed.  

            1 point: Arguments attempt to address key concepts, some gaps in 
            logic or comprehension.  Some use of scientific terms, majority 
            used correctly.

            0 points: Arguments are difficult to follow, multiple flaws in 
            logic.  Incorrect or no use of key terms. Connections between 
            concepts not present.

            The student uses data from the graphs to support their explanations.

            1 point: Yes, the writer uses data to support their explanation.

            0 points: No, the writer does not use sufficient data to support 
            their explanation.

            \textbf{Questions/Answers:}

            \textit{\{As Figure \ref{figure:zero-shot-cot-with-answers}.\}}

            \textbf{Grade: 10/10}

            - - -

            \textbf{Student's Answers:}

            \textit{\{As Figure \ref{figure:zero-shot-cot-with-answers}.\}}


            \textbf{Grade:}

            - - -

            \highlight{10/10}
            \vspace{-0.30cm}
            \highlight{The student correctly identified both detection methods and provided clear explanations of how each} 
            \vspace{-0.30cm}
            \highlight{ method works, earning 4 points. They also correctly identified physical characteristics for both}
            \vspace{-0.30cm}
            \highlight{exoplanets, earning 2 points. The student clearly identified exoplanet A as Earth-like, earning 1 point.}
            \vspace{-0.30cm}
            \highlight{The student's arguments were presented in a clear, logical way that demonstrated understanding of}
            \vspace{-0.30cm}
            \highlight{concepts, earning 2 points. Lastly, the student used data from the graphs  to support their explanations,} 
            \vspace{-0.30cm}
            \highlight{earning 1 point. No points were deducted as the student met all the criteria in the rubric.}

            };
        \end{tikzpicture}
    \end{minipage}
    \caption{A representation of the Zero-shot-CoT prompt that incorporates both instructor-provided correct answers and grading rubrics for grading students' assignments. The grading process utilized in the Zero-shot-CoT with correct answers only (Figure \ref{figure:zero-shot-cot-with-answers}) is also applied here.}
    \label{figure:zero-shot-cot-with-answers-and-rubrics}
\end{figure*}

\begin{figure*}[!]
    \begin{minipage}{\textwidth} 
        \centering
        \begin{tikzpicture}[rounded corners=8pt, thick, text=black, text opacity=1]
            \node[draw=solid_gray, fill=light_gray, line width=1pt, 
            text=black, text width=0.97\textwidth, align=left, 
            font=\fontsize{6pt}{1em}\selectfont, inner xsep=5pt, 
            inner ysep=5pt] at (0,0) {\textbf{Instruction:}
            Your task is to design a rubric that addresses the following questions/answers. 
            The followings are the homework assignments for the Astrobiology course. 
            This course is designed for undergraduate students majoring in Astronomy. 
            The rubric must be unbiased and adaptable, capable of fairly evaluating any 
            kind of student writing assignment.

            The scoring breakdown for each question should be as follows: the score for 
            Question 1 is 10, the score for Question 2 is 10, and the score for Question 3 
            is 10. Your thoughtful consideration in creating this rubric will 
            ensure all students' work is evaluated equitably and consistently. Make sure
            that the rubric will provide points that are whole numbered.

            Each rubric should be dedicated to each question separately.

            - - -
            
            \textbf{Questions/Answers:}

            QUESTION 1

            \textit{\{As Figure \ref{figure:zero-shot-cot-with-answers}.\}}

            ANSWER 1

            \textit{\{As Figure \ref{figure:zero-shot-cot-with-answers}.\}}

            Full Score: 10/10
            
            QUESTION 2

            Discuss how habitable zone range and spectral type are related. [...]

            \textit{\{The question is partially cut due to space constraints.\}}

            ANSWER 2

            The habitable zone is defined as the range of distances from a star within which liquid water can exist on a planet. [...]

            \textit{\{The answer is partially cut due to space constraints.\}}

            Full Score: 10/10

            QUESTION 3

            Discuss why/whether it is plausible that exobiology exists at all, 
            in terms of exoplanet type, spectral type and orbital distance. [...]

            \textit{\{The question is partially cut due to space constraints.\}}

            ANSWER 3

            The star around HD-127688 has a K spectral type.  [...]
            
           \textit{\{The answer is partially cut due to space constraints.\}}
            

            Full Score: 10/10

            - - -

            \highlight{Rubric for Question 1:}
            \vspace{-0.30cm}

            \highlight{- Identification of detection methods (2 points): The student correctly identifies the detection methods}
            \vspace{-0.30cm}
            \highlight{used for each exoplanet.}
            \vspace{-0.30cm}
            \highlight{- Explanation of detection methods (2 points): The student provides a clear and accurate explanation}
            \vspace{-0.30cm}
            \highlight{of how each detection method works.}
            \vspace{-0.30cm}
            \highlight{- Identification of physical characteristics (2 points): The student correctly identifies the physical} 
            \vspace{-0.30cm}
            \highlight{characteristics that can be learned from each set of data.}
            \vspace{-0.30cm}
            \highlight{- Explanation of physical characteristics (2 points): The student provides a clear and accurate} 
            \vspace{-0.30cm}
            \highlight{explanation of why these physical characteristics can be learned from the data.}
            \vspace{-0.30cm}
            \highlight{- Identification of Earth-like exoplanet (2 points): The student correctly identifies one exoplanet as} 
            \vspace{-0.30cm}
            \highlight{Earth-like.}
            \vspace{-0.30cm}

            \highlight{Rubric for Question 2:}
            \vspace{-0.30cm}

            \highlight{- Discussion of habitable zone and spectral type (2 points): The student provides a clear and accurate} 
            \vspace{-0.30cm}
            \highlight{explanation of how the habitable zone range and spectral type are related.}
            \vspace{-0.30cm}
            \highlight{- Response to Yousef's statement (2 points): The student clearly states whether they agree or disagree} 
            \vspace{-0.30cm}
            \highlight{with Yousef's statement and provides evidence and data to support their answer.}
            \vspace{-0.30cm}
            \highlight{- Response to Lora's statement (2 points): The student clearly states whether they agree or disagree} 
            \vspace{-0.30cm}
            \highlight{with Lora's statement and provides evidence and data to support their answer.}
            \vspace{-0.30cm}
            \highlight{- Use of data (2 points): The student effectively uses data to support their answers.}
            \vspace{-0.30cm}
            \highlight{- Clarity of explanation (2 points): The student's explanations are clear and logical. [...]}



            \textit{\{The answer is partially cut due to space constraints.\}}

            };
        \end{tikzpicture}
    \end{minipage}
    \caption{A depiction of the prompt employed to generate a rubric using GPT-4 for the Astrobiology course. We repeat this process for all the courses under study by replacing the course name, correct answer, total grade, and the question. The generated rubric is then embedded into the prompt containing Zero-shot-CoT and the correct answer for grading students' writing assignments. Specifically, the prompt displayed in Figure \ref{figure:zero-shot-cot-with-answers-and-rubrics} is used for grading, where the instructor-supplied rubric is substituted with an LLM-produced rubric.}
    \label{figure:template-prompt-for-LLM-generated-rubrics}
\end{figure*}

\noindent \textbf{Zero-shot-CoT with Instructor-provided Answers and Rubrics.}
For our second prompting technique, we build upon the previously mentioned prompt by incorporating the rubrics presented by the instructors for each question. This integration seeks to guide the LLM towards the grading methodology typically employed by instructors, thereby achieving a higher extent of alignment with the grades allocated by the instructors.
Retaining all components from the prior prompting technique, we include instructor-provided rubrics in the prompt and follow the same grading procedure as in the previous stage.
Figure \ref{figure:zero-shot-cot-with-answers-and-rubrics} illustrates the enhancement of the Zero-shot-CoT prompt with the inclusion of the correct answers and rubrics.

\noindent \textbf{Zero-shot-CoT with Instructor-provided Answers and LLM-generated Rubrics.}
Within this approach, we diverge from the other two previous methods where the reference information in the prompt is centered around the human (instructor) and instead, include a portion of information generated by the LLM. This idea is inspired by the fact that an LLM can generate better rubrics compared to those provided by a human, which can better align with human preferences. The reasoning behind this is that LLMs are trained on an extensive amount of data, and thus, possess comprehensive interdisciplinary knowledge, such as knowledge from both education and astronomy in our case, that allows them to produce improved rubrics. To implement this, we prompt GPT-4 (see Section \ref{section:experimental-setup} for more details) and utilize the template prompt shown in Figure \ref{figure:template-prompt-for-LLM-generated-rubrics} to generate rubrics based on the provided correct answer, total grade, and the question itself. Once the rubric is generated, it replaces the rubric used in the previous stage with the instructor-provided rubric (Figure \ref{figure:zero-shot-cot-with-answers-and-rubrics}), and the same process is repeated to grade all students' assignments.

\subsection{Bootstrap Resampling}
As previously discussed, MOOCs cater to a large number of participants for each course, and our study aims to evaluate the effectiveness of LLMs in grading assignments for a substantial student population. However, due to budget limitations of our project when using proprietary LLMs, such as GPT-4, as well as the need for careful analysis of the grades assigned by LLMs to individual questions, we conduct the grading process using LLMs on a more manageable subset of the entire collection of students' assignments (refer to Section \ref{section:experimental-setup} for further details). On the other hand, it remains crucial to establish the performance of LLMs in grading a significant volume of assignments, representative of the regular situations in MOOCs. To address this requirement based on the manageable subset of assignments we examine, we use the bootstrap resampling technique \cite{10.1214/aos/1176344552,EfroTibs93}.
Ultimately, we evaluate the performance of LLMs in assigning grades compared to those given by the instructors by analyzing the average grades generated per question in each course.

\subsection{Baseline}

We analyze average scores produced by the LLMs for each question across all courses, comparing them to the peer grading as the baseline and the instructor-assigned grades as the ground truth. By evaluating the average scores with respect to the ground truth, our goal is to identify the most effective type of prompt and its corresponding LLM. In this context, a prompt, along with its associated LLM, is deemed superior if it demonstrates greater alignment with the instructor-assigned grades compared to those obtained through peer grading.

\section{Experimental Setup}
\label{section:experimental-setup}


\noindent \textbf{Data.}
This study employs data gathered from three MOOCs: Introductory Astronomy, Astrobiology, and the History and Philosophy of Astronomy \cite{astrobiology,astronomy,HPA}.
The process of obtaining the data followed a strict ethical roadmap, guided by an Institutional Review Board from our university. The participants willingly filled out a survey and formally consented to allow their course data to be used for research. All data related to the students was de-identified and purposefully sampled to represent the full range of assignment grades.

Although we did our best to avoid bias in the sampling process, we are aware that our courses might not be universally representative of the global population, due to the inherent limitations in subsampling from the entire population. They are, however, representative of the audience on Coursera as a whole.  The majority of students in these MOOCs are between the ages of 25 and 44. A small majority of students are male in all enrolled courses with the smallest difference between gender in the Astrobiology course (45\% female versus 54\% male) and the greatest difference in gender in the History and Philosophy course (34\% female versus 65\% male). Geographically, the United States and India are the homes of most students. Finally, since all data are fully de-identified, the assignment selection is minimally influenced by gender or geography.

Regarding the attributes of the assignments, students are instructed to compose their answers using no more than 500 words. Nevertheless, this constraint on the submission length is not strictly imposed. Specifically, the Introductory Astronomy course comprises five assignments, each worth 9 points except for the first question, which carries 6 points. The Astrobiology course features three questions, each valued at 10 points, while the Philosophy and History of Astronomy course contains four questions, each with 4 points.





In all of our experiments within the paper, we select a subset consisting of 10 writing assignments submitted by students for every course. 
Moreover, for implementing bootstrap resampling, we resample 10,000 samples.
It is worth noting that the data used in this research was sourced from a proprietary repository, automatically factoring in the issue of data contamination in LLMs \cite{golchin2023time,2023arXiv231106233G}.

\noindent \textbf{Setting.} In all of our experiments involving GPT-4 and GPT-3.5, we utilize the \texttt{gpt-4-0613} and \texttt{gpt-3.5-turbo-0613} snapshots, which are accessed through the OpenAI API. We set the temperature to zero and the maximum token limit to 2048, tasking LLMs with generating grades for each question across all three courses. All mentioned hyperparameters are maintained consistently throughout the evaluation of all prompts and across all LLMs in our experiments.

\noindent \textbf{Instructor Grading.} 
In our study, we consider the grades assigned by instructors (university professors) to be our ground truth. These professors are subject matter experts in the fields relevant to each course, and meticulously evaluate the assignments under consideration in this study. The grading process by the instructors is conducted in a double-blind manner, i.e., the instructors do not hold any knowledge regarding peer grading or the students' personal information.




\noindent \textbf{Peer Grading.} 
For each writing assignment, three to four fellow classmates are randomly selected to evaluate the work using a grading rubric provided by the course instructor. This rubric is identical to the one utilized by the instructor for grading. Although the grading process is conducted anonymously to ensure impartiality, the final grade for each assignment is determined by calculating the median of the scores received from the peer graders.

\noindent \textbf{LLM-generated Rubrics.} We utilize GPT-4 to generate rubrics for questions (assignments) in each course using the template prompt shown in Figure \ref{figure:template-prompt-for-LLM-generated-rubrics}. These rubrics are generated based on the specific questions for each course, the respective correct answers supplied by the instructors, and the total score for each question. After the rubrics are generated, they are integrated into the template prompt containing Zer-shot-CoT and correct answers for grading, as displayed in Figure \ref{figure:zero-shot-cot-with-answers-and-rubrics}. It is worth mentioning that all the rubrics produced by GPT-4 are entirely independent of any rubrics provided by the instructors.

\begin{table}[!t]
    \centering
    \caption{Average grades from grading 10 students' writing assignments using the three proposed prompting techniques: Zero-shot-CoT with instructor-provided answers (Instr. Ans.), Zero-shot-CoT combined with both instructor-supplied answers and rubrics (Instr. Ans. \& Instr. Rub.), and Zero-shot-CoT incorporating the instructor-provided answers and LLM-generated rubrics (Instr. Ans. \& LLM Rub.). The results encompass three MOOC subjects: Introductory Astronomy (Astr.), Astrobiology (A. Bio.), and the History and Philosophy of Astronomy (H.P.A.). For each prompt, we employ both GPT-3.5 and GPT-4 to generate grades. The generated grades are compared to instructor-assigned grades (instr. grades) and peer grades.}
    \label{tab:raw-resullts}
    \small
    \renewcommand{\arraystretch}{1.2}

    \begin{adjustbox}{width=\textwidth,center}

        \begin{tabular}{@{}cccccccccc@{}}
            \toprule
                                                        &                                 &                                                                                        & \multicolumn{1}{l}{}                                                                & \multicolumn{3}{c}{\textbf{GPT-3.5}}                                                                                                                                                                                                                  & \multicolumn{3}{c}{\textbf{GPT-4}}                                                                                                                                                                                               \\ \cmidrule(lr){5-7}  \cmidrule(l){8-10}
            \multicolumn{1}{c|}{\textbf{Courses}}         & \multicolumn{1}{c|}{\textbf{Q}} & \multicolumn{1}{c|}{\textbf{\begin{tabular}[c]{@{}c@{}}Instr. \\ Grades\end{tabular}}} & \multicolumn{1}{c}{\textbf{\begin{tabular}[c]{@{}c@{}}Peer\\ Grades\end{tabular}}} \vrule width 1pt & \multicolumn{1}{c|}{\textbf{\begin{tabular}[c]{@{}c@{}}Instr. \\ Ans.\end{tabular}}} & \multicolumn{1}{c|}{\textbf{\begin{tabular}[c]{@{}c@{}}Instr.\\ Ans. \&\\ Instr. Rub.\end{tabular}}} & \multicolumn{1}{c}{\textbf{\begin{tabular}[c]{@{}c@{}}Instr.\\ Ans. \&\\ LLM Rub.\end{tabular}}} \vrule width 1pt & \multicolumn{1}{c|}{\textbf{\begin{tabular}[c]{@{}c@{}}Instr. \\ Ans.\end{tabular}}} & \multicolumn{1}{c|}{\textbf{\begin{tabular}[c]{@{}c@{}}Instr.\\ Ans. \&\\ Instr. Rub.\end{tabular}}} & \multicolumn{1}{c}{\textbf{\begin{tabular}[c]{@{}c@{}}Instr.\\ Ans. \&\\ LLM Rub.\end{tabular}}} \\ \midrule
            \multicolumn{1}{c|}{\multirow{5}{*}{Astr.}} & \multicolumn{1}{c|}{Q1}         & \multicolumn{1}{c|}{3.90}                                                              & \multicolumn{1}{c}{5.15} \vrule width 1pt                                                           & \multicolumn{1}{c|}{4.50}                                        & \multicolumn{1}{c|}{2.90}                                                        & \multicolumn{1}{c}{4.10} \vrule width 1pt                                                                          & \multicolumn{1}{c|}{4.75}                                       & \multicolumn{1}{c|}{4.40}                                                        & \multicolumn{1}{c}{4.40}                                                     \\ \cmidrule(l){2-10} 
            \multicolumn{1}{c|}{}                         & \multicolumn{1}{c|}{Q2}         & \multicolumn{1}{c|}{8.20}                                                              & \multicolumn{1}{c}{7.55} \vrule width 1pt                                                           & \multicolumn{1}{c|}{7.60}                                        & \multicolumn{1}{c|}{8.30}                                                        & \multicolumn{1}{c}{8.40} \vrule width 1pt                                                                          & \multicolumn{1}{c|}{8.65}                                       & \multicolumn{1}{c|}{8.30}                                                        & \multicolumn{1}{c}{8.50}                                                     \\ \cmidrule(l){2-10} 
            \multicolumn{1}{c|}{}                         & \multicolumn{1}{c|}{Q3}         & \multicolumn{1}{c|}{7.50}                                                               & \multicolumn{1}{c}{7.40} \vrule width 1pt                                                           & \multicolumn{1}{c|}{6.80}                                        & \multicolumn{1}{c|}{7.20}                                                        & \multicolumn{1}{c}{6.60} \vrule width 1pt                                                                          & \multicolumn{1}{c|}{7.60}                                        & \multicolumn{1}{c|}{7.30}                                                        & \multicolumn{1}{c}{7.60}                                                     \\ \cmidrule(l){2-10} 
            \multicolumn{1}{c|}{}                         & \multicolumn{1}{c|}{Q4}         & \multicolumn{1}{c|}{7.40}                                                               & \multicolumn{1}{c}{7.45} \vrule width 1pt                                                           & \multicolumn{1}{c|}{6.80}                                        & \multicolumn{1}{c|}{7.10}                                                        & \multicolumn{1}{c}{6.80} \vrule width 1pt                                                                          & \multicolumn{1}{c|}{7.50}                                        & \multicolumn{1}{c|}{6.90}                                                        & \multicolumn{1}{c}{7.05}                                                    \\ \cmidrule(l){2-10} 
            \multicolumn{1}{c|}{}                         & \multicolumn{1}{c|}{Q5}         & \multicolumn{1}{c|}{5.50}                                                               & \multicolumn{1}{c}{7.40} \vrule width 1pt                                                           & \multicolumn{1}{c|}{6.40}                                        & \multicolumn{1}{c|}{7.40}                                                        & \multicolumn{1}{c}{6.40} \vrule width 1pt                                                                          & \multicolumn{1}{c|}{6.20}                                        & \multicolumn{1}{c|}{5.90}                                                        & \multicolumn{1}{c}{6.35}                                                    \\ \midrule
            \multicolumn{1}{c|}{\multirow{3}{*}{A. Bio.}}   & \multicolumn{1}{c|}{Q1}         & \multicolumn{1}{c|}{6.80}                                                               & \multicolumn{1}{c}{7.50} \vrule width 1pt                                                            & \multicolumn{1}{c|}{7.40}                                        & \multicolumn{1}{c|}{6.70}                                                        & \multicolumn{1}{c}{6.58} \vrule width 1pt                                                                        & \multicolumn{1}{c|}{7.50}                                        & \multicolumn{1}{c|}{7.10}                                                        & \multicolumn{1}{c}{7.10}                                                     \\ \cmidrule(l){2-10} 
            \multicolumn{1}{c|}{}                         & \multicolumn{1}{c|}{Q2}         & \multicolumn{1}{c|}{6.70}                                                               & \multicolumn{1}{c}{7.45} \vrule width 1pt                                                           & \multicolumn{1}{c|}{7.00}                                          & \multicolumn{1}{c|}{7.30}                                                        & \multicolumn{1}{c}{5.52} \vrule width 1pt                                                                        & \multicolumn{1}{c|}{7.90}                                        & \multicolumn{1}{c|}{7.40}                                                        & \multicolumn{1}{c}{7.10}                                                     \\ \cmidrule(l){2-10} 
            \multicolumn{1}{c|}{}                         & \multicolumn{1}{c|}{Q3}         & \multicolumn{1}{c|}{7.90}                                                               & \multicolumn{1}{c}{9.05} \vrule width 1pt                                                           & \multicolumn{1}{c|}{6.70}                                        & \multicolumn{1}{c|}{6.50}                                                        & \multicolumn{1}{c}{5.02} \vrule width 1pt                                                                        & \multicolumn{1}{c|}{8.10}                                        & \multicolumn{1}{c|}{7.50}                                                        & \multicolumn{1}{c}{7.50}                                                     \\ \midrule
            \multicolumn{1}{c|}{\multirow{4}{*}{H.P.A.}}     & \multicolumn{1}{c|}{Q1}         & \multicolumn{1}{c|}{3.50}                                                               & \multicolumn{1}{c}{3.60} \vrule width 1pt                                                            & \multicolumn{1}{c|}{2.70}                                        & \multicolumn{1}{c|}{2.00}                                                          & \multicolumn{1}{c}{2.85} \vrule width 1pt                                                                         & \multicolumn{1}{c|}{3.50}                                        & \multicolumn{1}{c|}{3.20}                                                        & \multicolumn{1}{c}{3.20}                                                     \\ \cmidrule(l){2-10} 
            \multicolumn{1}{c|}{}                         & \multicolumn{1}{c|}{Q2}         & \multicolumn{1}{c|}{2.40}                                                               & \multicolumn{1}{c}{3.70} \vrule width 1pt                                                            & \multicolumn{1}{c|}{2.90}                                        & \multicolumn{1}{c|}{1.80}                                                        & \multicolumn{1}{c}{2.56} \vrule width 1pt                                                                        & \multicolumn{1}{c|}{3.25}                                       & \multicolumn{1}{c|}{3.10}                                                        & \multicolumn{1}{c}{2.95}                                                    \\ \cmidrule(l){2-10} 
            \multicolumn{1}{c|}{}                         & \multicolumn{1}{c|}{Q3}         & \multicolumn{1}{c|}{2.70}                                                               & \multicolumn{1}{c}{3.40} \vrule width 1pt                                                            & \multicolumn{1}{c|}{2.70}                                        & \multicolumn{1}{c|}{1.20}                                                        & \multicolumn{1}{c}{1.70} \vrule width 1pt                                                                          & \multicolumn{1}{c|}{3.65}                                       & \multicolumn{1}{c|}{3.20}                                                        & \multicolumn{1}{c}{3.20}                                                     \\ \cmidrule(l){2-10} 
            \multicolumn{1}{c|}{}                         & \multicolumn{1}{c|}{Q4}         & \multicolumn{1}{c|}{2.20}                                                               & \multicolumn{1}{c}{3.80} \vrule width 1pt                                                            & \multicolumn{1}{c|}{3.00}                                          & \multicolumn{1}{c|}{1.10}                                                        & \multicolumn{1}{c}{2.20} \vrule width 1pt                                                                          & \multicolumn{1}{c|}{3.25}                                       & \multicolumn{1}{c|}{2.70}                                                        & \multicolumn{1}{c}{2.95}                                                                         \\ \bottomrule
        \end{tabular}
    \end{adjustbox}
\end{table}


\section{Results and Discussion}
Detailed in Table \ref{tab:raw-resullts} and Table \ref{tab:bootstrap-results} are the outcomes of our evaluation of students' writing assignments using two LLMs—GPT-4 and GPT-3.5—based on a suite of three unique prompts crafted within this study. Table \ref{tab:raw-resullts} displays the grades directly obtained from grading the assignments of 10 students included in our research, while Table \ref{tab:bootstrap-results} showcases the results after the implementation of bootstrap resampling. 
Our examination unfolds several key insights:

\noindent \textbf{(1)} Both Table \ref{tab:raw-resullts} and Table \ref{tab:bootstrap-results} reveal that GPT-4 overall outperforms GPT-3.5 by generating grades that are more closely aligned with the grades provided by instructors, especially in the context of the History and Philosophy of Astronomy course. This remains constant across all three prompts assessed.

\noindent \textbf{(2)} Focusing on the prompts, ones that integrate both instructor-supplied rubrics and LLM-produced rubrics give rise to grades that are better aligned with grades given by instructors. This is true for both LLMs: GPT-4 and GPT-3.5.

\noindent \textbf{(3)} In almost all scenarios, GPT-4, when prompted with instructor-provided answers and rubrics, generates grades that are either superior to or on par with peer grades. This even includes the History and Philosophy of Astronomy course, which requires speculative/imaginative thinking abilities.

\noindent \textbf{(4)} As expected, the LLM-produced rubrics, on average, as displayed in Table \ref{tab:bootstrap-results}, can perform on the same level as the instructor-provided rubrics. This finding is interesting as it suggests that LLMs could be used for rubrics generation in educational settings with the purpose of automating the entire grading process, only requiring the provision of correct answers.

\noindent \textbf{(5)} From the results listed in both Table \ref{tab:raw-resullts} and Table \ref{tab:bootstrap-results}, it is clear that the most difficult course to grade is the History and Philosophy of Astronomy. Both LLMs and peer graders find it challenging to produce grades that are in close alignment with those assigned by instructors. However, even in these cases, the superior performance of GPT-4 remains apparent as its grades are closer to grades assigned by instructors compared to those allocated by peer grading.

\begin{table}[!t]
    \centering
        \caption{Average grades from grading 10 students' writing assignments based on bootstrap resampling with 10,000 resampled samples. Each average grade includes a subscripted standard deviation. Similar to Table \ref{tab:raw-resullts}, we employ three different prompting strategies: Zero-shot-CoT with instructor-provided answers (Instr. Ans.), Zero-shot-CoT combined with both instructor-presented answers and rubrics (Instr. Ans. \& Instr. Rub.), and Zero-shot-CoT incorporating the instructor-provided answers and LLM-generated rubrics (Instr. Ans. \& LLM Rub.), across three MOOC topics: Introductory Astronomy (Astr.), Astrobiology (A. Bio.), and the History and Philosophy of Astronomy (H.P.A.). For each experiment with each prompt, we utilize both GPT-3.5 and GPT-4. The generated grades are contrasted with the instructor-provided grades (instr. grades) and peer grades.}
    \label{tab:bootstrap-results}
    \small
    \renewcommand{\arraystretch}{1.2}
    \begin{adjustbox}{width=\textwidth,center}
        \begin{tabular}{@{}cccccccccc@{}}
            \toprule
                                                        &                                 &                                                                                        & \multicolumn{1}{l}{}                                                                & \multicolumn{3}{c}{\textbf{GPT-3.5}}                                                                                                                                                                                                                                                           & \multicolumn{3}{c}{\textbf{GPT-4}}                                                                                                                                                                                                                                        \\ \cmidrule(lr){5-7}  \cmidrule(l){8-10}
            \multicolumn{1}{c|}{\textbf{Courses}}         & \multicolumn{1}{c|}{\textbf{Q}} & \multicolumn{1}{c|}{\textbf{\begin{tabular}[c]{@{}c@{}}Instr. \\ Grades\end{tabular}}} & \multicolumn{1}{c}{\textbf{\begin{tabular}[c]{@{}c@{}}Peer\\ Grades\end{tabular}}} \vrule width 1pt & \multicolumn{1}{c|}{\textbf{\begin{tabular}[c]{@{}c@{}}Instr.\\ Ans.\end{tabular}}} & \multicolumn{1}{c|}{\textbf{\begin{tabular}[c]{@{}c@{}}Instr. \\ Ans. \&\\ Instr. Rub.\end{tabular}}} & \multicolumn{1}{c}{\textbf{\begin{tabular}[c]{@{}c@{}}Instr. \\ Ans. \&\\ LLM Rub.\end{tabular}}} \vrule width 1pt & \multicolumn{1}{c|}{\textbf{\begin{tabular}[c]{@{}c@{}}Instr.\\ Ans.\end{tabular}}} & \multicolumn{1}{c|}{\textbf{\begin{tabular}[c]{@{}c@{}}Instr. \\ Ans. \&\\ Instr. Rub.\end{tabular}}} & \textbf{\begin{tabular}[c]{@{}c@{}}Instr. \\ Ans. \&\\ LLM Rub.\end{tabular}} \\ \midrule
            \multicolumn{1}{c|}{\multirow{5}{*}{Astr.}}   & \multicolumn{1}{c|}{Q1}         & \multicolumn{1}{c|}{$3.90_{\pm0.54}$}                                                  & \multicolumn{1}{c}{$5.15_{\pm0.27}$} \vrule width 1pt                                               & \multicolumn{1}{c|}{$4.50_{\pm0.16}$}                                               & \multicolumn{1}{c|}{$2.90_{\pm0.61}$}                                                                & \multicolumn{1}{c}{$4.10_{\pm0.61}$} \vrule width 1pt                                                             & \multicolumn{1}{c|}{$4.75_{\pm0.41}$}                                               & \multicolumn{1}{c|}{$4.40_{\pm0.41}$}                                                                & $4.40_{\pm0.48}$                                                             \\ \cmidrule(l){2-10} 
            \multicolumn{1}{c|}{}                         & \multicolumn{1}{c|}{Q2}         & \multicolumn{1}{c|}{$8.20_{\pm0.37}$}                                                  & \multicolumn{1}{c}{$7.55_{\pm0.75}$} \vrule width 1pt                                               & \multicolumn{1}{c|}{$7.60_{\pm0.29}$}                                               & \multicolumn{1}{c|}{$8.30_{\pm0.25}$}                                                                & \multicolumn{1}{c}{$8.40_{\pm0.21}$} \vrule width 1pt                                                             & \multicolumn{1}{c|}{$8.65_{\pm0.20}$}                                               & \multicolumn{1}{c|}{$8.30_{\pm0.28}$}                                                                & $8.50_{\pm0.27}$                                                             \\ \cmidrule(l){2-10} 
            \multicolumn{1}{c|}{}                         & \multicolumn{1}{c|}{Q3}         & \multicolumn{1}{c|}{$7.51_{\pm0.92}$}                                                  & \multicolumn{1}{c}{$7.41_{\pm0.82}$} \vrule width 1pt                                               & \multicolumn{1}{c|}{$6.81_{\pm0.75}$}                                               & \multicolumn{1}{c|}{$7.21_{\pm0.85}$}                                                                & \multicolumn{1}{c}{$6.61_{\pm0.78}$} \vrule width 1pt                                                             & \multicolumn{1}{c|}{$7.61_{\pm0.87}$}                                               & \multicolumn{1}{c|}{$7.31_{\pm0.86}$}                                                                & $7.61_{\pm0.87}$                                                             \\ \cmidrule(l){2-10} 
            \multicolumn{1}{c|}{}                         & \multicolumn{1}{c|}{Q4}         & \multicolumn{1}{c|}{$7.41_{\pm0.86}$}                                                  & \multicolumn{1}{c}{$7.45_{\pm0.46}$} \vrule width 1pt                                               & \multicolumn{1}{c|}{$6.81_{\pm0.78}$}                                               & \multicolumn{1}{c|}{$7.11_{\pm0.91}$}                                                                & \multicolumn{1}{c}{$6.81_{\pm0.94}$} \vrule width 1pt                                                             & \multicolumn{1}{c|}{$7.51_{\pm0.77}$}                                               & \multicolumn{1}{c|}{$6.91_{\pm0.90}$}                                                                & $7.06_{\pm0.89}$                                                             \\ \cmidrule(l){2-10} 
            \multicolumn{1}{c|}{}                         & \multicolumn{1}{c|}{Q5}         & \multicolumn{1}{c|}{$5.51_{\pm0.94}$}                                                  & \multicolumn{1}{c}{$7.40_{\pm0.83}$} \vrule width 1pt                                               & \multicolumn{1}{c|}{$6.41_{\pm0.78}$}                                               & \multicolumn{1}{c|}{$7.40_{\pm0.91}$}                                                                & \multicolumn{1}{c}{$6.41_{\pm1.10}$} \vrule width 1pt                                                             & \multicolumn{1}{c|}{$6.21_{\pm0.90}$}                                               & \multicolumn{1}{c|}{$5.91_{\pm1.06}$}                                                                & $6.36_{\pm1.10}$                                                             \\ \midrule
            \multicolumn{1}{c|}{\multirow{3}{*}{A. Bio.}} & \multicolumn{1}{c|}{Q1}         & \multicolumn{1}{c|}{$6.81_{\pm1.09}$}                                                  & \multicolumn{1}{c}{$7.50_{\pm0.79}$} \vrule width 1pt                                               & \multicolumn{1}{c|}{$7.41_{\pm0.84}$}                                               & \multicolumn{1}{c|}{$6.71_{\pm1.18}$}                                                                & \multicolumn{1}{c}{$6.60_{\pm0.94}$} \vrule width 1pt                                                             & \multicolumn{1}{c|}{$7.50_{\pm0.70}$}                                               & \multicolumn{1}{c|}{$7.11_{\pm0.95}$}                                                                & $7.11_{\pm0.85}$                                                             \\ \cmidrule(l){2-10} 
            \multicolumn{1}{c|}{}                         & \multicolumn{1}{c|}{Q2}         & \multicolumn{1}{c|}{$6.71_{\pm0.83}$}                                                  & \multicolumn{1}{c}{$7.46_{\pm1.02}$} \vrule width 1pt                                               & \multicolumn{1}{c|}{$7.01_{\pm0.76}$}                                               & \multicolumn{1}{c|}{$7.31_{\pm0.83}$}                                                                & \multicolumn{1}{c}{$5.53_{\pm1.30}$} \vrule width 1pt                                                             & \multicolumn{1}{c|}{$7.91_{\pm0.68}$}                                               & \multicolumn{1}{c|}{$7.41_{\pm0.83}$}                                                                & $7.11_{\pm0.98}$                                                             \\ \cmidrule(l){2-10} 
            \multicolumn{1}{c|}{}                         & \multicolumn{1}{c|}{Q3}         & \multicolumn{1}{c|}{$7.89_{\pm0.85}$}                                                  & \multicolumn{1}{c}{$9.04_{\pm0.52}$} \vrule width 1pt                                               & \multicolumn{1}{c|}{$6.70_{\pm0.71}$}                                               & \multicolumn{1}{c|}{$6.49_{\pm0.85}$}                                                                & \multicolumn{1}{c}{$5.02_{\pm1.35}$} \vrule width 1pt                                                             & \multicolumn{1}{c|}{$8.10_{\pm0.33}$}                                               & \multicolumn{1}{c|}{$7.50_{\pm0.41}$}                                                                & $7.50_{\pm0.29}$                                                             \\ \midrule
            \multicolumn{1}{c|}{\multirow{4}{*}{H.P.A.}}     & \multicolumn{1}{c|}{Q1}         & \multicolumn{1}{c|}{$3.50_{\pm0.21}$}                                                  & \multicolumn{1}{c}{$3.60_{\pm0.21}$} \vrule width 1pt                                               & \multicolumn{1}{c|}{$2.70_{\pm0.14}$}                                               & \multicolumn{1}{c|}{$2.00_{\pm0.00}$}                                                                & \multicolumn{1}{c}{$2.85_{\pm0.20}$} \vrule width 1pt                                                             & \multicolumn{1}{c|}{$3.50_{\pm0.21}$}                                               & \multicolumn{1}{c|}{$3.20_{\pm0.31}$}                                                                & $3.20_{\pm0.22}$                                                             \\ \cmidrule(l){2-10} 
            \multicolumn{1}{c|}{}                         & \multicolumn{1}{c|}{Q2}         & \multicolumn{1}{c|}{$2.39_{\pm0.29}$}                                                  & \multicolumn{1}{c}{$3.69_{\pm0.20}$} \vrule width 1pt                                               & \multicolumn{1}{c|}{$2.90_{\pm0.22}$}                                               & \multicolumn{1}{c|}{$1.80_{\pm0.31}$}                                                                & \multicolumn{1}{c}{$2.56_{\pm0.25}$} \vrule width 1pt                                                             & \multicolumn{1}{c|}{$3.25_{\pm0.23}$}                                               & \multicolumn{1}{c|}{$3.10_{\pm0.17}$}                                                                & $2.95_{\pm0.23}$                                                             \\ \cmidrule(l){2-10} 
            \multicolumn{1}{c|}{}                         & \multicolumn{1}{c|}{Q3}         & \multicolumn{1}{c|}{$2.70_{\pm0.20}$}                                                  & \multicolumn{1}{c}{$3.40_{\pm0.38}$} \vrule width 1pt                                               & \multicolumn{1}{c|}{$2.69_{\pm0.32}$}                                               & \multicolumn{1}{c|}{$1.20_{\pm0.42}$}                                                                & \multicolumn{1}{c}{$1.70_{\pm0.38}$} \vrule width 1pt                                                             & \multicolumn{1}{c|}{$3.65_{\pm0.14}$}                                               & \multicolumn{1}{c|}{$3.20_{\pm0.19}$}                                                                & $3.20_{\pm0.18}$                                                             \\ \cmidrule(l){2-10} 
            \multicolumn{1}{c|}{}                         & \multicolumn{1}{c|}{Q4}         & \multicolumn{1}{c|}{$2.20_{\pm0.28}$}                                                  & \multicolumn{1}{c}{$3.80_{\pm0.19}$} \vrule width 1pt                                               & \multicolumn{1}{c|}{$3.00_{\pm0.14}$}                                               & \multicolumn{1}{c|}{$1.10_{\pm0.33}$}                                                                & \multicolumn{1}{c}{$2.20_{\pm0.19}$} \vrule width 1pt                                                             & \multicolumn{1}{c|}{$3.25_{\pm0.24}$}                                               & \multicolumn{1}{c|}{$2.70_{\pm0.28}$}                                                                & $2.95_{\pm0.23}$                                                             \\ \bottomrule
        \end{tabular}
    \end{adjustbox}
\end{table}

\section{Conclusion}
We introduced three distinct prompts based on the zero-shot chain-of-thought (Zero-shot-CoT) prompting technique for utilizing LLMs to automate the grading of students' assignments in educational platforms with substantial student populations, such as MOOCs. Specifically, our method consists of Zero-shot-CoT with instructor-supplied correct answers, Zero-shot-CoT augmented with instructor-offered answers and rubrics, and Zero-shot-CoT combined with correct answers provided by the instructors and LLM-generated rubrics.

We examined the performance of each proposed prompting strategy using two state-of-the-art LLMs—GPT-4 and GPT-3.5—by grading assignments from three MOOCs: Introductory Astronomy, Astrobiology, and the History and Philosophy of Astronomy. Our results suggested that the integration of Zero-shot-CoT with instructor-presented correct answers and rubrics provided by either the instructor or LLM, when using GPT-4 as the underlying foundation model, surpasses peer grading across all subjects. We also discovered that grading in courses demanding reasoning and imaginative thinking skills, such as the History and Philosophy of Astronomy, is more difficult in comparison to others.

In summary, we found LLMs more effective than peer grading for deriving grades that closely align with the ones assigned by instructors on high-traffic educational platforms.


%
%
%
\bibliographystyle{splncs04}
\bibliography{paper}

\end{document}